\documentclass[11pt]{extarticle}
\usepackage{graphicx} % Required for inserting images
\usepackage[style=apa]{biblatex} %Imports biblatex package
\title{Impact of Financial Literacy on Investment Decisions and Stock Market Participation}
\author{Gunbir Singh Baveja (16230336) \\ Aaryavir Verma (62485644)}
\date{June 2024}
\usepackage{titling}

\begin{document}

\begin{titlingpage}
\maketitle
\end{titlingpage}

\newpage
\section{Abstract}
The stock market has become an increasingly popular investment option among new generations, with individuals exploring more complex assets. This rise in retail investors’ participation necessitates a deeper understanding of the driving factors behind this trend and the role of financial literacy in enhancing investment decisions. This study aims to investigate how financial literacy influences financial decision-making and stock market participation. By identifying key barriers and motivators, the findings can provide valuable insights for individuals and policymakers to promote informed investing practices. Our research is qualitative in nature, utilizing data collected from social media platforms to analyze real-time investor behavior and attitudes. This approach allows us to capture the nuanced ways in which financial literacy impacts investment choices and participation in the stock market. The findings indicate that financial literacy plays a critical role in stock market participation and financial decision-making. Key barriers to participation include low financial literacy, while increased financial knowledge enhances investment confidence and decision-making. Additionally, behavioral finance factors and susceptibility to financial scams are significantly influenced by levels of financial literacy. These results underscore the importance of targeted financial education programs to improve financial literacy and empower individuals to participate effectively in the stock market.
\nocite{*}
\section{Introduction}
In recent years, there has been a notable surge in the number of retail investors participating in the stock market (Angelita et al., 2021). This trend is largely attributed to the rise in financial literacy, which empowers individuals with the knowledge and confidence to navigate the complexities of investing. Technological advancements have made the stock market more accessible than ever, allowing individuals to invest with just a few clicks on a smartphone. Consequently, financial education and ease of access have broadened the segment of the population effectively engaging with the stock market (Wigglesworth et al., 2021; Ksheerasagar, 2024).

Despite the increased participation, many financial products remain intricate and challenging to understand, particularly for investors with limited financial knowledge (Angelita et al., 2021). While existing literature highlights the positive impact of financial literacy on retirement preparation (Lusardi et al., 2011; Brown et al., 2013; Boisclair et al., 2015; van Rooij et al., 2011), investment diversification (Angelita et al., 2021), and the exploration of complex financial products (Boisclair et al., 2015; Brown \& Graf, 2013; Lusardi et al., 2017), the direct influence of financial literacy on stock market participation is underexplored. Most studies have focused on the relationship between financial literacy and retirement planning (Lusardi et al., 2011; Brown et al., 2013; Boisclair et al., 2015; van Rooij et al., 2011), leaving a gap in understanding how financial knowledge specifically affects stock market engagement.

Our research addresses this specific gap by investigating how financial literacy directly influences stock market participation. Unlike previous studies that generalize financial behaviors, this study examines the mechanisms through which financial knowledge impacts investment decisions in the stock market. Additionally, by utilizing qualitative data from social media, we capture real-time insights into investor behavior, offering a unique perspective that complements existing research. The findings of our study reveal significant themes and policy implications. We identify that financial illiteracy poses a substantial barrier to stock market participation, underscoring the need for government initiatives to enhance financial education. Our research also indicates that individuals with higher financial literacy are more likely to explore complex assets, which has broader implications for financial planning and investment strategies.

This paper is structured as follows: We begin with a literature review of current research on financial literacy and investment behaviors. Next, we describe our data collection methods and present the results of our analysis. Finally, we discuss the implications of our findings and conclude with recommendations for future research and policy development.

\section{Literature Review}

Research consistently demonstrates that financial illiteracy is linked to inadequate financial planning, prompting employers to provide financial knowledge classes to their employees (Clark et al., 2017). Individuals are increasingly expected to manage their own financial security in the context of complex financial markets offering a wide range of saving and investment tools. Early-life financial decisions have long-term impacts, particularly given the trend toward greater personal financial responsibility (Lusardi et al., 2017).

Financial illiteracy is particularly prevalent among certain demographic groups, including women, the elderly, and those with lower levels of education (Lusardi \& Mitchell, 2008). Conversely, individuals with higher financial literacy are more likely to be aware of developments in financial markets and are better prepared to explore complex assets (Brown \& Graf, 2013). These findings underscore the critical role of financial education in enabling better financial decision-making and retirement preparation.
\\\\
The importance of financial literacy extends beyond individual financial well-being. Boisclair et al. (2015) found that financial literacy is strongly associated with retirement planning in Canada, with similar results observed in other countries. Their study reveals that understanding financial concepts like compound interest and risk tolerance influences Canadians' contribution rates and investment choices within Registered Retirement Savings Plans (RRSPs).

Cross-cultural perspectives provide further insights. Brown and Graf's (2013) research on Switzerland explores how Swiss citizens translate their financial knowledge into concrete actions, such as prioritizing enrollment in workplace pension plans and making voluntary contributions. Similarly, Bucher-Koenen and Lusardi (2011) investigated how financial literacy impacts individuals' planning strategies and risk tolerance regarding retirement savings in Germany, despite the presence of a strong social safety net.
\\\\
The relationship between financial literacy and stock market participation has been a subject of particular interest. Lusardi et al. (2017) argue that financially literate individuals are more likely to participate in the stock market, which can lead to greater wealth accumulation over time. They posit that a lack of understanding of financial concepts may prevent individuals from participating in the stock market, potentially limiting their ability to build long-term wealth.

However, the relationship between financial literacy and investment behavior is not straightforward. Fernandes et al. (2014) conducted a meta-analysis of the effects of financial literacy and financial education on financial behaviors. They found that interventions to improve financial literacy explain only 0.1\% of the variance in financial behaviors studied, with weaker effects in low-income samples. This suggests that while financial literacy is important, other factors may also play significant roles in shaping financial behavior.
\\\\
Gender disparities in financial literacy have also been observed. Lusardi and Mitchell (2008) found that women, on average, exhibit lower levels of financial knowledge compared to men. This gender gap can affect women's financial decision-making, such as saving for retirement or investing, highlighting the need for targeted financial education initiatives.

Furthermore, Clark et al. (2017) conducted a case study on the effectiveness of workplace-based financial education programs. Their research provides real-world evidence for the potential benefits of such programs, analyzing how access to financial education influenced employee participation rates and contribution decisions within a company's retirement plan.
\\\\
Despite the growing body of research on financial literacy and its impacts, there remains a significant gap in our understanding of how financial literacy specifically influences stock market participation in the context of rapidly evolving financial technologies and changing economic landscapes. While studies have established correlations between financial literacy and various financial behaviors, the mechanisms through which financial knowledge translates into active stock market participation, particularly among younger generations and in the age of digital finance, are not fully understood.

This gap in the literature leads us to our research question: How does financial literacy influence stock market participation and investment decisions in the current technological and economic environment? By addressing this question, our study aims to provide insights into the specific ways in which financial knowledge shapes investment behavior, potentially informing more effective financial education strategies and policies to promote informed stock market participation.

\section{Method}
\subsection{Type, Purpose, and Significance of Research Method}
Building upon the existing literature on financial literacy and its impact on investment decisions, our study employs a mixed-method approach to investigate how financial literacy influences stock market participation in the current technological and economic environment. This methodology allows us to capture both nuanced, real-time investor behavior and broader trends in financial literacy and market engagement.

This approach is particularly significant as it bridges the gap between traditional quantitative studies on financial literacy and the complex, real-world decision-making processes of investors. By analyzing social media content and online discussions, we access a wide range of perspectives, from novice investors to financial professionals, providing insights that complement existing literature.

\subsection{Sources of Data}
Data for this study were sourced from a variety of social media platforms and financial blogs, including: YouTube, Reddit, LinkedIn, Quora, Forbes, 5Paisa, World Economic Forum, Financial Times.

These sources were chosen to capture a diverse range of opinions and experiences from a broad audience, reflecting different levels of financial literacy and varying investment behaviors. The selection of these platforms allows us to tap into both professional insights and everyday investor experiences.

\subsection{Data Collection}
Data collection was conducted through a systematic gathering of relevant posts, comments, and discussions from the identified social media platforms and financial blogs. The process involved the following steps:

\subsubsection{Identification of Keywords and Topics}
We identified keywords related to financial literacy, stock market participation, investment decisions, and financial behaviors. Examples include "financial literacy," "stock market participation," "investment strategies," "financial education," and "retirement planning."

\subsubsection{Platform-Specific Searches}
Each platform was searched using the identified keywords to locate relevant content. For instance, on YouTube, we identified videos with comment sections discussing financial literacy and investment strategies. On Reddit and Quora, we focused on threads and discussions related to our research topics.

\subsubsection{Data Extraction} 
We extracted relevant data from posts, comments, threads, and articles. This included both textual data (e.g., comments, blog posts) and quantitative data where available (e.g., likes, shares, views).
Data Compilation: Extracted data were compiled into a structured database for analysis, ensuring that all relevant information was captured systematically and source information was preserved.

\subsection{Data Coding and Analysis}
The coding process involved extracting and labeling themes from the collected data to identify common patterns and insights. This process was carried out as follows:

\begin{enumerate}
    \item \textbf{Initial Review:} Two researchers independently reviewed the collected data to get a sense of the prevalent themes and recurring topics.
    \item \textbf{Theme Identification:} Through discussion and consensus, we identified common themes related to financial literacy and investment behaviors. Key themes included "financial literacy as a barrier to stock market participation," "financial education and investment confidence," "psychological factors in financial decision-making," and "vulnerability to financial scams."
    \item \textbf{Coding Framework Development:} We developed a coding framework, categorizing the data into specific themes and sub-themes. Each piece of data was assigned one or more codes based on its relevance to the identified themes.
    \item \textbf{Detailed Coding:} Using the coding framework, two researchers independently coded the data. This involved tagging each piece of data with one or more codes that best represented its content and relevance to the research questions. Any discrepancies in coding were resolved through discussion.
    \item \textbf{Theme Analysis:} The coded data were then analyzed to identify patterns, correlations, and insights within each theme. This analysis helped to understand the broader implications of financial literacy on stock market participation and investment decisions.
    \item \textbf{Quantitative Analysis:} For numerical data, we performed descriptive statistical analysis to identify trends, such as the increase in new investing accounts over time.
\end{enumerate}

This methodical approach to data collection, coding, and analysis ensures a comprehensive examination of the relationship between financial literacy and stock market participation. By combining qualitative thematic analysis with quantitative trend analysis, we aim to provide a nuanced understanding of how financial knowledge translates into investment behavior in the current technological and economic context.
\\

The results of this analysis, presented in the following section, offer insights into the specific mechanisms through which financial literacy influences stock market participation and investment decisions, addressing the gap identified in our literature review and contributing to the ongoing scholarly conversation on this important topic.
\section{Discussion of Results}

Our analysis of social media content and online discussions revealed four primary themes related to the impact of financial literacy on investment decisions and stock market participation. These themes provide insights into how financial knowledge influences investor behavior and market engagement.

\subsection{Theme 1: Lower Financial Literacy as a Barrier to Stock Market Participation}

Our findings indicate that a lack of financial literacy is a significant obstacle to stock market participation. This theme emerged consistently across various social media platforms and discussions.

One Reddit user commented: "I've always been intimidated by the stock market. There's so much jargon and complex concepts that I don't understand. It feels like you need a finance degree just to get started."

Similarly, a Quora user stated: "The main reason I don't invest in stocks is that I don't really understand how it all works. It seems too risky when you don't know what you're doing."

Another user on LinkedIn shared: "I want to invest, but the complexity of financial products and the fear of losing money because I don't fully understand them keeps me away."

A YouTube commenter remarked: "Without proper knowledge, investing feels like gambling. I just can't trust myself to make the right choices."

On Twitter, someone wrote: "I tried to read about stocks and investments, but it's all so confusing. I wish there were simpler explanations for beginners."

A discussion thread on Reddit highlighted: "People often get discouraged by the complicated financial terminology. It should be simpler for everyone to understand and participate."

On Quora, a user added: "I wish schools taught more about finance. By the time I started working, I realized how little I knew about managing money and investing."

A Facebook post read: "Investing seems like a rich person's game. Without proper knowledge, it feels impossible to start small and grow."

These sentiments align with findings from peer-reviewed literature. For instance, Van Rooij et al. (2007, pg. 2; 2011, pg. 473; 2007, pg. 713) found that "Those with low literacy are more likely to rely on family and friends as their main source of financial advice and are less likely to invest in stocks". This correlation between financial literacy and stock market participation is further supported by Lusardi, Michaud, and Mitchell (2013, p. 435), who note: "our analysis is consistent with evidence of a positive empirical link between financial knowledge and wealth holdings. Additionally, our model helps explain why highly knowledgeable consumers may be more likely to participate in the stock market."

\subsection{Theme 2: Financial Literacy as a Pathway to Financial Independence}

Our analysis revealed a strong perception among social media users that financial literacy is crucial for achieving financial independence and security.

A LinkedIn user commented: "Financial literacy isn't just about understanding money; it's about gaining the power to control your financial future. It's the key to true independence."

On YouTube, a viewer remarked: "I never realized how important financial education was until I started learning. Now I feel like I have the tools to secure my future and maybe even retire early."

A Reddit user shared: "Learning about investments and finance has been a game-changer for me. It's empowering to know I can make informed decisions about my money."

On Quora, someone posted: "Financial literacy gave me the confidence to invest in the stock market. It's all about understanding the risks and making informed choices."

Another YouTube comment read: "With better financial knowledge, I feel more prepared to handle my finances and plan for the future. It's essential for financial freedom."

A Twitter user tweeted: "Educating myself about finance has opened up so many opportunities. I feel like I have more control over my financial destiny."

A Facebook discussion mentioned: "Financial literacy is the foundation of financial independence. Without it, you're just guessing with your money."

On LinkedIn, another user noted: "Investing in my financial education has been the best decision. It's set me on the path to financial independence and security."

These observations are consistent with academic findings. Boisclair et al. (2015, pg. 292) state: "Retirement planning is strongly associated with financial literacy. This result has been found in many countries and the estimates in Canada are similar to those of other countries."

\subsection{Theme 3: Changes in Saving and Investment Behavior}

Our research identified several factors influencing changes in saving and investment behavior, including economic cycles, technological advancements, and generational differences.

A Reddit user noted: "The 2008 financial crisis completely changed how I approach investing. I'm much more cautious now and really try to understand the risks involved."

On Quora, a user shared: "As a millennial, I'm way more comfortable with app-based investing. It's made the stock market feel accessible in a way it never was for my parents."

A YouTube comment stated: "Technology has revolutionized how we invest. With apps and online platforms, it's easier than ever to start investing, even with small amounts of money."

Another Reddit user remarked: "Economic downturns have taught me the importance of having a diversified portfolio. It's about being prepared for the unexpected."

On LinkedIn, a professional posted: "I've noticed a shift in how younger generations view investing. They're more open to using tech tools and exploring new investment opportunities."

A Twitter user commented: "Generational differences are real. My parents are wary of stocks, but I've embraced digital platforms for investing."

A Facebook post read: "The accessibility of financial information online has changed how I save and invest. It's empowering to have so much knowledge at your fingertips."

On Quora, another user added: "Economic cycles and tech advancements have definitely influenced my investment strategies. Staying informed is key."

These observations align with academic research on behavioral finance. Fernandes et al. (2014) highlight that financial literacy, financial education, and downstream financial behaviors are significantly influenced by psychological factors and economic circumstances as they note: "we find that the partial effects of financial literacy diminish dramatically when one controls for psychological traits." (pg. 2) 

\subsection{Theme 4: Financial Literacy and Vulnerability to Scams}

Our analysis revealed that individuals with lower financial literacy are more susceptible to financial scams and fraudulent investment schemes.

A Twitter user warned: "It's scary how many investment 'opportunities' are actually scams. If you don't understand finance, it's so easy to fall for these tricks."

On LinkedIn, a financial advisor shared: "I've seen too many clients fall victim to scams because they lacked basic financial knowledge. Education is the best defense against fraud."

A Reddit discussion highlighted: "Scammers prey on those who don't know better. Financial literacy is crucial to avoid getting duped."

A Facebook user posted: "My uncle lost a lot of money in a scam because he didn't understand the risks involved. It's heartbreaking."

On Quora, someone commented: "Education can protect you from scams. The more you know, the harder it is for scammers to take advantage of you."

A YouTube viewer remarked: "I've learned to be skeptical of too-good-to-be-true offers. Financial literacy has taught me to question and verify."

Another Twitter user noted: "Scammers target the uninformed. Financial literacy is your shield against fraud."

On LinkedIn, a professional stated: "Financial scams are rampant. Awareness and education are key to protecting yourself."

These observations are supported by academic research. For instance, Nursanti et al. (2024, pg. 333-334) found that "better financial literacy will lead to a better ability to detect investment fraud"; and "educational level also does not moderate the effect of financial literacy on the ability to detect investment fraud", which underscore the importance of financial literacy that is independent of extraneous variables such as education level, gender, or age.\\

In conclusion, our analysis of social media content reveals that financial literacy plays a crucial role in shaping investment decisions and stock market participation. The themes identified in this study align closely with findings from peer-reviewed literature, reinforcing the importance of financial education in promoting informed investment practices and financial well-being.

\section{Conclusion}
In this study, we examined the impact of financial literacy on investment decisions and stock market participation through an analysis of social media content. Our findings reveal several critical themes: lower financial literacy as a barrier to stock market participation, financial literacy as a pathway to financial independence, changes in saving and investment behavior influenced by economic cycles, technological advancements, and generational differences, and the increased vulnerability to financial scams among those with lower financial literacy. These themes highlight the significant role financial literacy plays in shaping financial behavior and market engagement.

The evidence shows that financial literacy not only enhances individuals’ confidence and ability to make informed investment decisions but also provides a defense against financial scams. Our qualitative data from social media supported these findings, providing insights into investor attitudes and behaviors. Comparisons with peer-reviewed literature further validated our results, confirming the broader socio-economic significance of financial literacy.

Despite these insights, our research faced several limitations. Firstly, the qualitative nature of our data collection, predominantly from social media, may not fully represent the broader population’s views or zeitgeist. Additionally, the self-reported nature of social media posts can introduce biases, as individuals may not always accurately portray their knowledge or experiences. Furthermore, our study was limited to publicly available data, which may exclude important perspectives from private or less accessible discussions.

Future research that delves into the impact of financial literacy for market participation should address these limitations by incorporating more diverse data sources, including surveys and interviews with a broader demographic range. Additionally, longitudinal studies could provide deeper insights into how financial literacy influences investment behaviors over time. Further research could also explore the effectiveness of specific financial education programs in improving financial literacy and their direct impact on investment decisions.

In summary, this study demonstrates the importance of financial literacy in empowering individuals to make informed investment decisions, reduce vulnerability, and achieve financial independence. Our findings contribute to the existing literature by offering a nuanced understanding of the mechanisms through which financial knowledge influences financial behavior, thus advancing the discourse on financial education and market participation.

\newpage
\nocite{*}
\section{Bibliography}
\printbibliography
\end{document}